\documentclass[10pt,twocolumn,letterpaper]{article}

\usepackage{cvpr}              
\definecolor{cvprblue}{rgb}{0.21,0.49,0.74}
\usepackage[pagebackref,breaklinks,colorlinks,allcolors=cvprblue]{hyperref}

\usepackage{xcolor}
\usepackage[table]{xcolor}

\usepackage[numbers]{natbib}

\usepackage{bm}
\usepackage{color}
\usepackage{xcolor}
\usepackage{multirow}
\usepackage{ifthen}
\usepackage{dsfont}
\usepackage{pifont}
\usepackage{booktabs}
\usepackage{algorithm}
\usepackage{algorithmic}
\usepackage{graphicx}

\usepackage{nicefrac}
\usepackage{bigstrut}

\usepackage{float}

\usepackage{tcolorbox}
\tcbset{
  colframe=gray!75, colback=white, coltitle=black,
  sharp corners,
  boxrule=0.5mm,
  fonttitle=\bfseries,
  coltext=black,
  left=5pt, right=5pt, top=5pt, bottom=5pt
}

\usepackage{listings}
\tcbuselibrary{listings}

\usepackage{fancyvrb}
\usepackage{tcolorbox}
\tcbuselibrary{breakable}

\newcommand{\hk}[1]{#1}

\definecolor{cvprblue}{rgb}{0.21,0.49,0.74}
\usepackage[pagebackref,breaklinks,colorlinks,allcolors=cvprblue]{hyperref}

\definecolor{myred}{RGB}{200,0,0}
\newcommand{\myparagraph}[1]{\noindent\textbf{#1}}

\def\paperID{3695} 
\def\confName{CVPR}
\def\confYear{2026}
\title{EgoGraph: Temporal Knowledge Graph for Egocentric Video Understanding}

\author{Shitong Sun\\
Queen Mary University of London\\
\and
Ke Han\\
University of Trento\\
\and
Yukai Huang\\
\and
Weitong Cai\\
Queen Mary University of London\\
\and
Jifei Song\\
University of Surrey
}

\begin{document}
\maketitle

 
\begin{figure*}[!t]
   \centering
   \includegraphics[width=\textwidth]{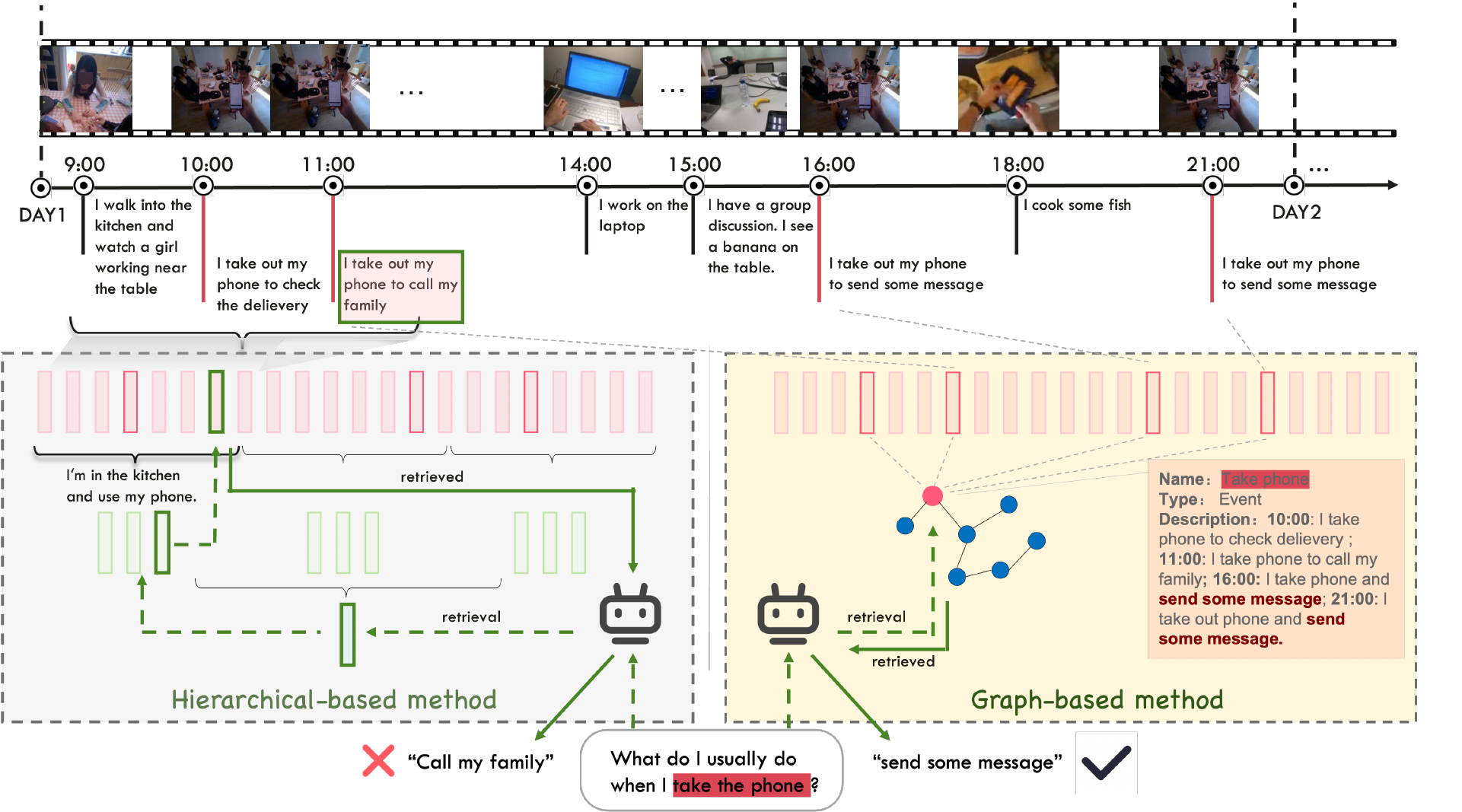}

   \caption{
   Comparison between hierarchical and graph-based methods for ultra-long egocentric video understanding. The hierarchical method summarizes each clip-level segment independently, whereas our graph-based approach constructs an entity-centric memory that models long-term dependencies across temporally separated events.
   \label{fig:motivation}} \vspace{-0.3cm}
\end{figure*}

\begin{abstract}

\hk{
Ultra-long egocentric videos spanning multiple days present significant challenges for video understanding. Existing approaches still rely on fragmented local processing and limited temporal modeling, restricting their ability to reason over such extended sequences.
To address these limitations, we introduce EgoGraph, a training-free and dynamic knowledge-graph construction framework that explicitly encodes long-term, cross-entity dependencies in egocentric video streams. 
EgoGraph employs a novel egocentric schema that unifies the extraction and abstraction of core entities, such as people, objects, locations, and events, and structurally reasons about their attributes and interactions, yielding a significantly richer and more coherent semantic representation than traditional clip-based video models.
Crucially, we develop a temporal relational modeling strategy that captures temporal dependencies across entities and accumulates stable long-term memory over multiple days, enabling complex temporal reasoning.
Extensive experiments on the EgoLifeQA and EgoR1-bench benchmarks demonstrate that EgoGraph achieves state-of-the-art performance on long-term video question answering, validating its effectiveness as a new paradigm for ultra-long egocentric video understanding.}

\end{abstract}    
\section{Introduction}
\label{sec:intro}


Daily activity logging is becoming increasingly essential for both humans and embodied agents, driven by the widespread adoption of wearable cameras in augmented reality devices and robotic platforms. These egocentric videos provide continuous first-person visual streams that naturally capture daily experiences, playing a crucial role in downstream tasks such as episodic memory retrieval and question answering~\cite{grauman2022ego4d,datta2022episodic}.
Despite significant progress in video understanding, existing methods primarily focus on short clips of less than one hour~\cite{jin2024chat,li2023videochat}, 
 and consequently struggle to generalize to \hk{real-world long-term scenarios where relevant information is embedded within day-long contexts \cite{yang2025egolife}}. 
This challenge of ultra-long video understanding is particularly crucial for egocentric videos, where the core difficulty lies not merely in recognizing what happens, but in capturing when events occur and how they temporally relate across extended time spans.

\hk{
However, research on this topic remains very limited, with \cite{yang2025egolife} representing the pioneering attempt to tackle ultra-long egocentric video understanding.
This work segments long videos into short clips and converts them into textual captions. The captions are then organized hierarchically through multi-level summarization (hourly, daily), leveraging LLMs' strong textual reasoning capabilities for retrieval.
Despite its effectiveness, this paradigm exhibits two fundamental limitations.
1) It processes short video clips separately, overlooking inter-clip dependencies and long-range temporal dynamics. Consequently, semantically related events that occur across distant time spans are fragmented into disconnected textual pieces, making it difficult to reason about temporal relationships among people, actions, and events.
2) Enormous yet fragmented captions are continuously produced, resulting in a large-scale but unstructured information space that potentially limits the model’s scalability and retrieval efficiency.
}
These limitations motivate an alternative representation: 
rather than hierarchically aggregating information, a 
structured knowledge graph can explicitly preserve entity 
relationships and temporal dependencies. As illustrated in 
Figure~\ref{fig:motivation}, graph-based representation 
maintains fine-grained connections that are lost in 
hierarchical summarization.

\hk{Building on this insight, we introduce EgoGraph, a structured temporal knowledge graph that dynamically stores and evolves over time to represent condensed egocentric information for long-term video understanding.
Our key insight is to frame a human-like structured memory based on events, which analyzes inter-entity relationships at specific moments and explicitly connects temporal relationships of entities across time through consistent inference over video streams.
Unlike existing static graphs in other domains (e.g., knowledge inference, course understanding \cite{ren2025videorag,chu2025understanding,rodin2024action}), which struggle to capture the dynamic, person- and event-driven nature of daily life, EgoGraph is specifically designed for egocentric videos that involve long-term, multi-person interactions from a first-person viewpoint.}

\hk{To construct structured graphs, we design an egocentric schema that defines node types for key semantic elements in egocentric videos, such as persons, events, objects, and their associated attributes, analogous to maintaining condensed personal profiles and event records. This schema enables a comprehensive yet informative representation of egocentric video content.
More importantly, EgoGraph consistently models temporal connections across multiple levels, including nodes, edges, and graph chunks, and performs reasoning over the graph, such as inferring a person’s habits from multiple observations, tracking changes in an object’s location.
It further forms relational hubs that connect participating entities with their temporal context while maintaining the graph compact through node incorporation.
This design also allows temporally constrained queries to retrieve relevant subgraphs directly, rather than scanning the entire video history, thereby enhancing retrieval efficiency.
Consequently, video contexts are incrementally stored and updated within a structured and compact graph, substantially improving both the efficiency and accuracy of video understanding.
}

\hk{
Our main contributions are as follows:
(1) We introduce EgoGraph, a training-free temporal knowledge graph framework for ultra-long egocentric video understanding, effectively overcoming the fragmented processing and limited temporal modeling of existing approaches.
(2) We propose an egocentric schema that constructs structured entities and cross-entity relationships, together with a temporal relational modeling strategy that captures long-range dependencies across days, enabling efficient and coherent long-term reasoning.
(3) We conduct extensive experiments on ultra-long egocentric benchmarks EgoLife and EgoR1-bench, demonstrating that EgoGraph achieves state-of-the-art performance on video question answering, significantly outperforming existing models.
}

\section{Related Work}
\label{sec:related_work}
\noindent\textbf{Egocentric Video Understanding.}
Egocentric videos captured from wearable cameras record the first-person experiences of daily activities, presenting unique challenges for video understanding due to their long-form nature, frequent viewpoint changes, and dense object interactions~\cite{grauman2022ego4d}. The Ego4D benchmark~\cite{grauman2022ego4d} established large-scale tasks for egocentric video analysis, with episodic memory retrieval as a core challenge—requiring systems to answer queries such as ``Where did I last see X?" by localizing relevant temporal windows in past video streams~\cite{datta2022episodic}. Recent advances leverage multi-modal large language models (MLLMs) for egocentric question answering~\cite{wang2023lifelongmemory,zhang2024multifactor}, demonstrating the potential of vision-language models for this domain. However, these approaches face fundamental limitations when processing long videos: EgoTempo~\cite{plizzari2025omnia} reveals that current models often rely on static frames or commonsense reasoning rather than temporal dynamics, while QaEgo4D~\cite{barmann2022keys} highlights the difficulty of maintaining constant-sized memory representations for extended video sequences. Action Scene Graphs~\cite{rodin2024action} introduce graph-based representations to capture evolving object relationships, yet remain limited to static structural modeling without explicit temporal attributes. These challenges motivate structured representations that can simultaneously capture relational knowledge and temporal dynamics inherent to egocentric video understanding.

\noindent\textbf{Graph-based Visual Understanding.}
Recent methods introduce structured knowledge graphs to capture relational information in videos. Scene graphs~\cite{johnson2015image} represent objects as nodes and their spatial or semantic relationships as edges, enabling explicit reasoning over visual content. Early work applied scene graphs to video captioning~\cite{pan2020spatio} and visual question answering~\cite{teney2017graph}, demonstrating improved performance through structured relational modeling. Dynamic scene graph generation methods~\cite{cong2021spatial,nag2023unbiased} extend this paradigm to videos by constructing frame-level graphs and modeling temporal evolution of relationships across adjacent frames. Recent advances integrate knowledge graphs with video retrieval: VideoRAG~\cite{ren2025videorag} constructs static knowledge graphs for lecture videos, while GraphVideoAgent~\cite{chu2025understanding} tracks object appearance status to build temporal graphs. However, these approaches face critical limitations for egocentric video understanding. VideoRAG's static graphs cannot capture the dynamic, event-driven nature of first-person experiences, while GraphVideoAgent's object-centric tracking lacks global context to inter-connect entities through shared events. More fundamentally, existing methods treat temporal information either implicitly through sequential construction or as localized object trajectories, rather than embedding time as first-class graph attributes with filtering capabilities. This gap becomes critical for episodic memory queries that demand knowing not just what happened, but precisely when events occurred and how they temporally relate.













\begin{figure*}[t]
   
   \centering
   \includegraphics[width=\textwidth]{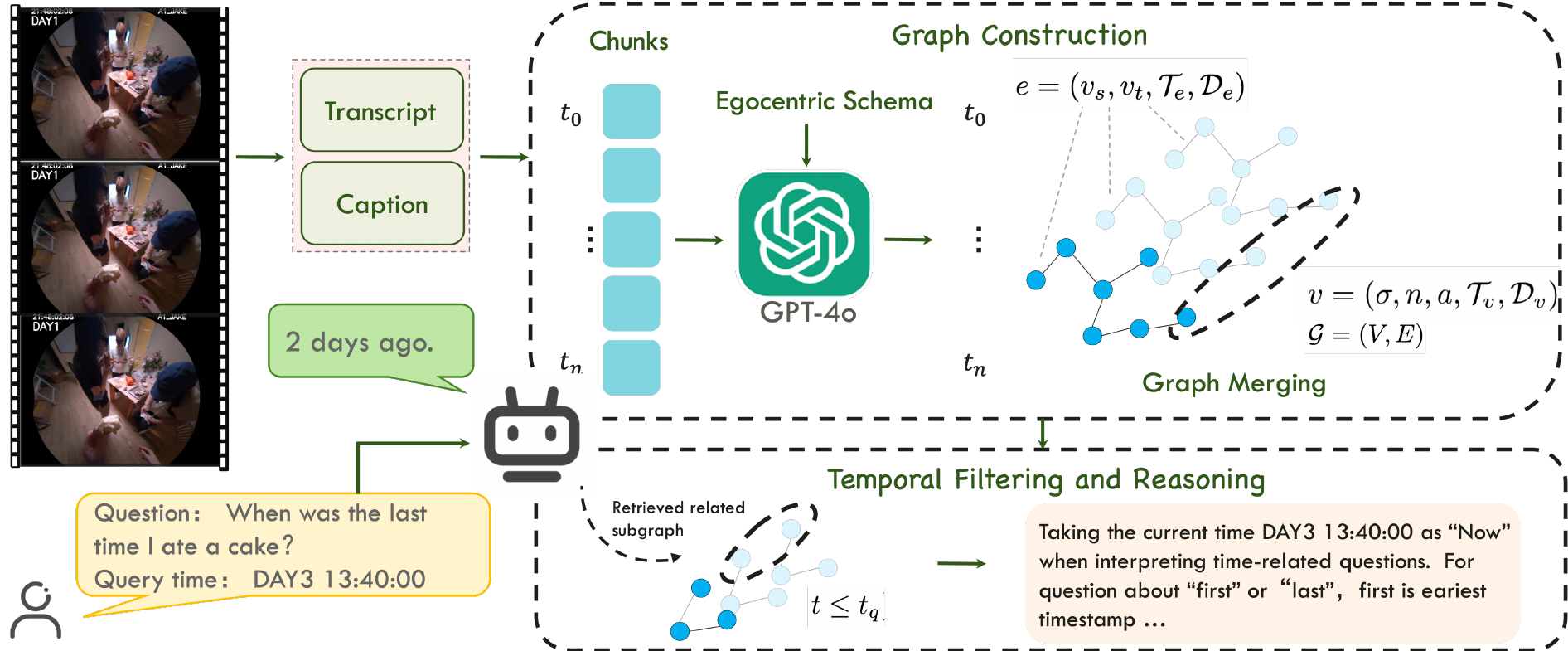}

   \caption{The pipeline of our proposed EgoGraph. EgoGraph encodes ultra-long video into a temporal-aware event knowledge graph, which imitates human brain memory processing. 
   }\vspace{-0.5cm}

   \label{fig:framework}
\end{figure*}

\section{Methodology}

Current methods for egocentric video understanding suffer from fragmented information processing and insufficient temporal modeling in ultra-long videos.
To address these limitations, we propose EgoGraph, a structured temporal knowledge graph 
that explicitly encodes long-term dependencies among entities, to improve memory efficiency and video understanding accuracy.
We first 
formalize EgoGraph as a temporal knowledge graph and present 
its construction pipeline from egocentric videos (Section~\ref{sec:TKG_define}). 
Then, we introduce our question answering framework that 
leverages EgoGraph for temporal reasoning and answer generation 
(Section ~\ref{sec:qa}).

\subsection{Temporal-aware EgoGraph}
\label{sec:TKG_define}

To enable temporal reasoning over long-form egocentric videos, 
we propose EgoGraph, a temporal knowledge graph that captures 
entities, relations, and their temporal dynamics from first-person 
observations. This section presents the formal definition and 
construction pipeline of EgoGraph.
%
%
We define the temporal-aware EgoGraph as:
\begin{equation}
\mathcal{G} = (V, E),
\end{equation}
where each entity $v \in V$ is represented as:
\begin{equation}
\label{eq_node}
v = (\sigma, n, a, \mathcal{T}_v, \mathcal{D}_v),
\end{equation}
where $\sigma$ denotes entity type;
$n$ is entity name; $a$ represent the predefined attributes for the entity type $\sigma$; $ \mathcal{T}_v, \mathcal{D}_v $ are a set of timestamps and descriptions, respectively.
Each edge $e \in E$, represented as 
\begin{equation}
\label{eq_edge}
e = (v_s, v_t, \mathcal{T}_e, \mathcal{D}_e) , 
\end{equation}
connects entities $v_s$ and $v_t$ through relation description $d \in \mathcal{D}$ at timestamps $t \in \mathcal{T}$.

Given an egocentric video $\mathcal{V}$ spanning multiple days, our goal is to construct a temporal knowledge graph $\mathcal{G}$. For a query $q$ at timestamp $t_q$, we retrieve answers from the temporally-filtered graph:
\begin{equation}
\mathcal{G}^{\leq t_q} = \{(v, e) \in \mathcal{G} : \forall t \in \mathcal{T}_v \cup \mathcal{T}_e, t \leq t_q\}.
\end{equation}

\subsubsection{EgoGraph Construction}
\hk{
For temporal modeling in graph construction, we begin by establishing temporal grounding for video content descriptions $\mathcal{S}$, which include dense captions and video transcripts, following prior works \cite{yang2025egolife}.
To process the potentially unbounded length of descriptions, we partition descriptions into chunks $\{c_1, c_2, \ldots, c_M\}$ subject to a maximum token constraint $L_{\text{max}}$.
Critically, each chunk is labeled with a timestamp $t_i$, which indicates its temporal anchor from the earliest timestamp within its span.
This anchoring strategy ensures that extracted entities are attributed to their first observed occurrence, maintaining temporal causality in downstream graph construction.}


\hk
{For each temporally anchored chunk $(c_i, t_i)$, we leverage large language models (LLMs) to extract entities and relations for graph construction.
Prior graph-based methods often impose few constraints on entity types, which can result in trivial or redundant semantics in egocentric scenarios, thereby reducing graph effectiveness.
To address this, we propose an egocentric schema that guides the extraction of nontrivial semantics and enforces semantic consistency across the entire graph.}


\myparagraph{Egocentric Schema.}
\hk{Personal episodic memory typically involves the who, where, and what of past experiences, e.g., the people encountered, locations visited, and contexts of events.
Inspired by how the human brain organizes memory, our proposed egocentric schema captures the essential entity types and their attributes in egocentric scenarios.
We define four core entity types, i.e., \texttt{Person}, \texttt{Location}, \texttt{Object}, and \texttt{Event}, each associated with specific attributes, as summarized in Table~\ref{tab:entity_types}.
This attribute-enriched representation ensures that the resulting graph is not merely a collection of mentions but a semantically rich and structurally organized knowledge abstraction.}

\hk{Moreover, this schema ensures semantic consistency through well-defined entity types that enable type-aware reasoning, captures the subject–object–location–action interaction patterns characteristic of egocentric videos, and maintains scalability by preventing unbounded growth of entity types in long-term video streams.}
%
%
%

\begin{table}[t]
\centering
\caption{Entity Type Definitions with Attributes}
\label{tab:entity_types}
\small
\setlength{\tabcolsep}{4pt}
\begin{tabular}{@{}lp{0.35\linewidth}p{0.45\linewidth}@{}}
\toprule
\textbf{Type} & \textbf{Attributes} & \textbf{Example} \\
\midrule
PERSON   & name, gender, appearance, preferences, dislikes, habits, hometown & 
         \texttt{Person("John", gender="male", hometown="London")} \\
\addlinespace
LOCATION & name, description & 
         \texttt{Location("kitchen", desc="office kitchen area")} \\
\addlinespace
OBJECT   & name, type, color, size, condition, owner, purchase\_information & 
         \texttt{Object("coffee\_mug", color="yellow", owner="John")} \\
\addlinespace
EVENT    & name, description, start\_time, subject, object, location & 
         \texttt{Event("meeting", loc="kitchen", subject="John")} \\
\bottomrule
\end{tabular}
\vspace{-0.5cm}
\end{table}


\myparagraph{Temporal Modeling and Reasoning.}
\hk{
To enable temporal modeling over ultra-long videos, we associate each entity and relation with a list of timestamps indicating when they were observed. Each timestamp records both the day and the time of day, allowing the system to track entity evolution and retrieve historical context at any point along the video timeline.
As formulated in Eq.~(\ref{eq_node}) and (\ref{eq_edge}), $\mathcal{T}_v$ or $\mathcal{T}_e = {t_1, t_2, \ldots, t_n}$ denotes an ordered list of timestamps, each following the format \texttt{[DAY\textit{d} HH:MM:SS]}, where \textit{d} represents the recording day and \texttt{HH:MM:SS} specifies the time within that day.
Whenever a new timestamp $t_i$ is added to the graph, new temporal connections can be established, triggering temporal reasoning in which the LLM infers higher-level conclusions such as personal habits, interpersonal relationships, and person–object interactions.
For instance, if the nodes “Jack” and “playing the piano” are repeatedly connected across multiple days, the model can infer that Jack enjoys playing the piano.
}

\hk{
For timestamped queries, our method can use the query timestamp as a temporal reference point, which not only localizes the relevant temporal context but also filters out future time steps, thereby improving reasoning efficiency.
}

\myparagraph{EgoGraph Merging and Update.}
\hk{
A fundamental challenge in graph construction lies in the rapid accumulation of redundant nodes over time, which expands the graph scale while diminishing its efficiency.
Unlike existing methods that either discard temporal information through aggregation or create duplicate nodes for each observation~\cite{ren2025videorag}, our temporal graph: 1) continuously merges redundant nodes based on their textual embedding similarity, and 2) appends new temporal information ($\mathcal{T}$ and $\mathcal{D}$) to preserve the complete evolutionary trajectory, where
\begin{equation}
\mathcal{T} = \left\{t_1, t_2, \ldots, t_n\right\}, \quad
\mathcal{D} = \left\{d_1, d_2, \ldots, d_n\right\},
\end{equation}
and each timestamp $t_i$ is paired with a corresponding description $d_i$ that records new information about node states or edge connections.
For example, attributes gathered from the same entity across different time points are consolidated into a single unified node, where the most recent non-empty attribute values are updated and existing values are retained.}
\hk{This design enables the graph to dynamically capture long-term dependencies among entities and relations in a scalable and semantically consistent way, rather than expanding the graph scale through redundant node accumulation.}



\subsection{Question Answering with EgoGraph}
\label{sec:qa}

\hk{
Question answering represents an important application of video understanding.
Given a question $q$ with an associated timestamp $t_q$ about the egocentric video, and following the retrieval-augmented generation process~\cite{guo2024lightrag}, we leverage the constructed graph to retrieve relevant information for answering the question.
Specifically, we extract semantic keywords from the query $q$ and perform vector similarity search over the entity embeddings to retrieve the top-$k$ most relevant entities.
In parallel, we retrieve the top-$k$ related edges and combine their contextual information to form a comprehensive reasoning context.}

\noindent
\textbf{Temporal Filtering.}
\hk{Since questions can be raised at any point in the video stream, we further define a temporal filtering operation to improve answering efficiency by extracting a subgraph that includes only temporally aligned information:
\begin{equation}
\mathcal{G}(t \leq t_q) = (V^{\leq t_q}, E^{\leq t_q}),
\end{equation}
where $V^{\leq t_q}$ and $E^{\leq t_q}$ denote the nodes and edges that satisfy the temporal constraint $t \leq t_q$.
This operation filters entities and their associated information based on the query timestamp, ensuring temporally coherent reasoning.}

\noindent
\textbf{LLM-based Temporal Reasoning.}
\label{sec:reasoning}
Given retrieved entities and relations with their timestamps, we leverage the 
temporal reasoning capabilities of LLMs through structured 
prompting.
The LLM performs temporal reasoning by comparing timestamps against the query 
reference point and applying the provided temporal rules. It receives the retrieved context along with explicit 
temporal reasoning instructions that define how to interpret time-related 
expressions relative to query timestamp $t_q$. The prompt specifies that 
all timestamps follow format \texttt{[DAY\textit{d} HH:MM:SS]} where 
$d$ indicates the date, with total ordering by day then time-of-day, and 
establishes $t_q$ as the reference point (``NOW'') for all temporal 
interpretation. Relative time expressions are resolved through rules 
provided in the prompt: ``yesterday'' maps to $\text{day}(t_q) - 1$, 
``last time'' selects $\max\{t \in \tau \mid t < t_q\}$, ``first time'' 
selects $\min\{t \in \tau\}$, and expressions like ``2 hours ago'' compute 
$t_q - \text{2h}$. The prompt explicitly requires timestamp citations 
for all referenced events, enabling answer verification and temporal 
grounding. This approach handles diverse 
temporal expressions in natural language without exhaustive rule enumeration, 
leveraging the LLM's language understanding while maintaining temporal 
accuracy through explicit instruction and structured output requirements.



\begin{table*}[t]
\centering
\caption{Performance comparison of EgoGraph with state-of-the-art models on EgoLifeQA and EgoR1-bench benchmarks. }
\resizebox{\textwidth}{!}{
\begin{tabular}{l c c c c c c c}
\toprule 
\multirow{2}{*}{\textbf{Model}}& \multicolumn{6}{c}{\textbf{EgoLifeQA}} &\multirow{2}{*}{\textbf{EgoR1-Bench}}\\
\cmidrule(lr){2-7} 
 &  \textbf{EntityLog} & \textbf{EventRecall} & \textbf{HabitInsight} & \textbf{RelationMap} & \textbf{TaskMaster} & \textbf{Average} & \\
\hline
\multicolumn{8}{c}{\textit{MLLMs}} \\
\hline
Gemini-1.5-Pro~\cite{team2023gemini}  & 36.0 & 37.3 & \textbf{45.9} & 30.4 & 34.9 & 36.9 &  38.3\\
GPT-4o~\cite{hurst2024gpt}  &  34.4 & 42.1 & 29.5 & 30.4 & 44.4 &   36.2 & - \\
LLaVA-OV~\cite{li2024llava} & 36.8 & 34.9 & 31.1 & 22.4 & 28.6 & 30.8  &  31.6\\
EgoGPT~\cite{yang2025egolife} (EgoIT-99K+D1) & 39.2 & 36.5 & 31.1 & 33.6 & 39.7 & 36.0  & - \\
EgoGPT~\cite{yang2025egolife} (InternVL-3.5-8B)  & 28.0 & 31.7 & 31.1 & 34.4 & 38.1 & 32.2 & 31.3 \\

\hline
\multicolumn{8}{c}{\textit{Graph-based methods}} \\
\hline
LightRAG~\cite{guo2024lightrag}  &40.8  &  40.4 & 36.0 & 32.0 & 50.7 & 39.2 & 31.1 \\
VideoRAG~\cite{ren2025videorag}  & 19.2 & 15.9 & 19.7 & 28.0 & 14.3 & 20.0 & 19.8 \\
\hline
EgoGraph & \textbf{46.4} & \textbf{46.8} & \textbf{45.9} & \textbf{35.2} & \textbf{60.3} & \textbf{45.8} &  \textbf{41.3} \\
\bottomrule 
\end{tabular}}

\label{tab:main}
\end{table*}

\section{Experiments}

\subsection{Datasets and Evaluation Metrics}
We adopt zero-shot evaluation to test on the following two datasets:
\noindent\textbf{EgoLifeQA}~\cite{yang2025egolife}
 constructed a super-long egocentric video
dataset in which six individuals wore camera-equipped glasses to capture their daily experiences
over seven days in a shared house, recording their daily activities such as cooking, shopping, and socializing. It publishes 500 public question-answering pairs for evaluating ultra-long video question answering.
\noindent\textbf{EgoR1-Bench}~\cite{tian2025ego} is
a reasoning based benchmark for ultra-long egocentric video
understanding, which comprises 300 QAs evenly distributed
across six first-person perspectives.



\subsection{Implementation Details}

We leverage \texttt{InternVL-3.5-8B-Instruct}~\cite{wang2025internvl3_5} as the visual captioner to extract visual information from videos.
The long videos are segmented into two-minute clips, with each clip processed alongside its corresponding transcript as input to the captioner.
For graph construction, we employ \texttt{gpt-4o} to perform entity and relation extraction within our designed constraints, and use the same model for question answering.
To retrieve relevant nodes and edges, we implement \texttt{text-embedding-3-small} from OpenAI as the text encoder and calculate cosine similarity to identify the top-$k$ nodes and edges, where $k=40$.
Each entity is stored with its description at each timestamp, which are automatically summarized when the number of timestamps exceeds 100 to maintain conciseness.

\subsection{Comparison with State-of-the-Arts}

\noindent\textbf{Baselines.} We compare EgoGraph with state-of-the-art methods for ultra-long video understanding:
(1)~\textbf{EgoGPT}~\cite{yang2025egolife} constructs hierarchical memory banks over long-term egocentric videos and leverages retrieval-augmented generation for question answering. It generates summaries based on video captions at both hour-level and day-level granularities, and retrieves temporal information through a top-down process from day to specific hours.
(2)~\textbf{VideoRAG}~\cite{ren2025videorag} integrates static knowledge graphs with videos. It is designed for lecture videos, where visual variation is significantly less pronounced compared to egocentric scenarios, in which head movements lead to substantial viewpoint changes.
(3)~\textbf{LightRAG}~\cite{guo2024lightrag} is a text-based method from NLP that first segments long contexts into chunks and then extracts nodes and edges from each chunk.
This graph-based retrieval-augmented generation approach is particularly suitable for query-focused summarization~\cite{zhang2024comprehensive}. Our scenario shares a similar philosophy, as we track the same entities appearing across temporally separated segments of ultra-long video contexts.
Note that all existing methods lack explicit temporal modeling and are not suitable for ultra-long egocentric video scenarios, which EgoGraph addresses through temporal knowledge graph sequences with explicit temporal reasoning.
For fair comparison, we employ InternVL-3.5-8B as the captioner for all methods unless otherwise specified. We also implement the same question answering model, \texttt{gpt-4o}, across all compared methods.


\noindent\textbf{Results.}
We provide a comprehensive evaluation of EgoGraph against state-of-the-art (SOTA) methods on the EgoLifeQA and EgoR1-Bench benchmarks, with detailed results presented in Table~\ref{tab:main}. Our competitors include leading MLLMs (e.g., Gemini-1.5-Pro, GPT-4o) and prior graph-based approaches (e.g., LightRAG, VideoRAG). On the primary EgoLifeQA benchmark, EgoGraph achieves a new SOTA average accuracy of 45.8\%, significantly surpassing all other methods. This represents a substantial lead of +6.6 points over the best-performing graph-based competitor, LightRAG (39.2\%), and +8.9 points over the strongest MLLM, Gemini-1.5-Pro (36.9\%). This superiority is consistent across all five sub-categories: EgoGraph achieves the highest or tied-for-highest score in every single task. Notably, it establishes a commanding lead in complex reasoning tasks such as TaskMaster (60.3\%) and EventRecall (46.8\%), while also matching the powerful Gemini-1.5-Pro in HabitInsight (45.9\%).

Furthermore, we validate our model's robustness on the EgoR1-Bench. EgoGraph again achieves the top score of 41.3\%, outperforming the formidable Gemini-1.5-Pro (38.3\%) by a clear margin of +3.0 points. While powerful MLLMs like Gemini-1.5-Pro are strong generalists, our results demonstrate their limitations in specialized, long-term egocentric reasoning compared to a dedicated, structured approach. The poor performance of VideoRAG (20.0\%) also suggests that naive graph construction is insufficient for this challenge. In contrast, EgoGraph's consistent top-tier performance across both benchmarks unequivocally establishes its superior capability to structure, retrieve, and reason over complex, long-duration egocentric video.

\subsection{Ablation Study}

\subsubsection{Hierarchical vs Graph}
\begin{figure}[ht]
   \centering
   \includegraphics[width=0.7\linewidth]{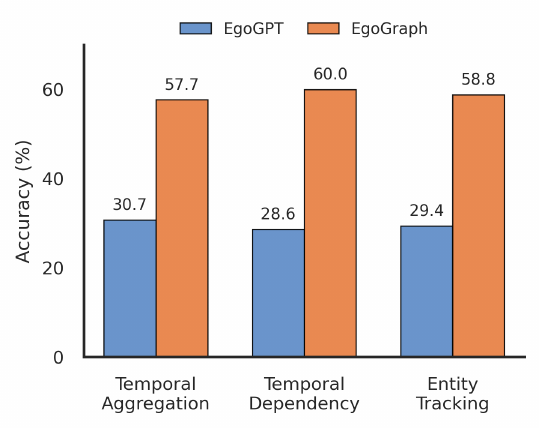}
   \vspace{-0.3cm}
   \caption{Quantitative comparison on temporal reasoning tasks. EgoGraph outperforms EgoGPT across temporal aggregation, temporal dependency, and entity tracking, achieving an average improvement of 29.3\%.
   \label{fig:ablation1}\vspace{-0.3cm}}
   
\end{figure}

We compare hierarchical-based methods with our graph-based approach on a series of subsets from the EgoLife dataset. We extract three subsets from the EgoLifeQA dataset based on temporal characteristics: (1) temporal aggregation, which includes questions containing ``usually"; (2) temporal dependency, which includes questions containing ``after"; and (3) entity tracking, which includes questions containing ``where". We focus on these question types because they involve entities that are temporally separated across long contexts.
As shown in Figure~\ref{fig:ablation1}, EgoGraph substantially outperforms EgoGPT across all three categories, achieving an average improvement of 29.3\%. This strongly supports our claim that graph-based methods with instance-centric representations, compared to hierarchical methods with continuous temporal information, can better capture long-term dependencies.

\begin{figure*}[ht]
   \centering
\includegraphics[width=0.9\linewidth]{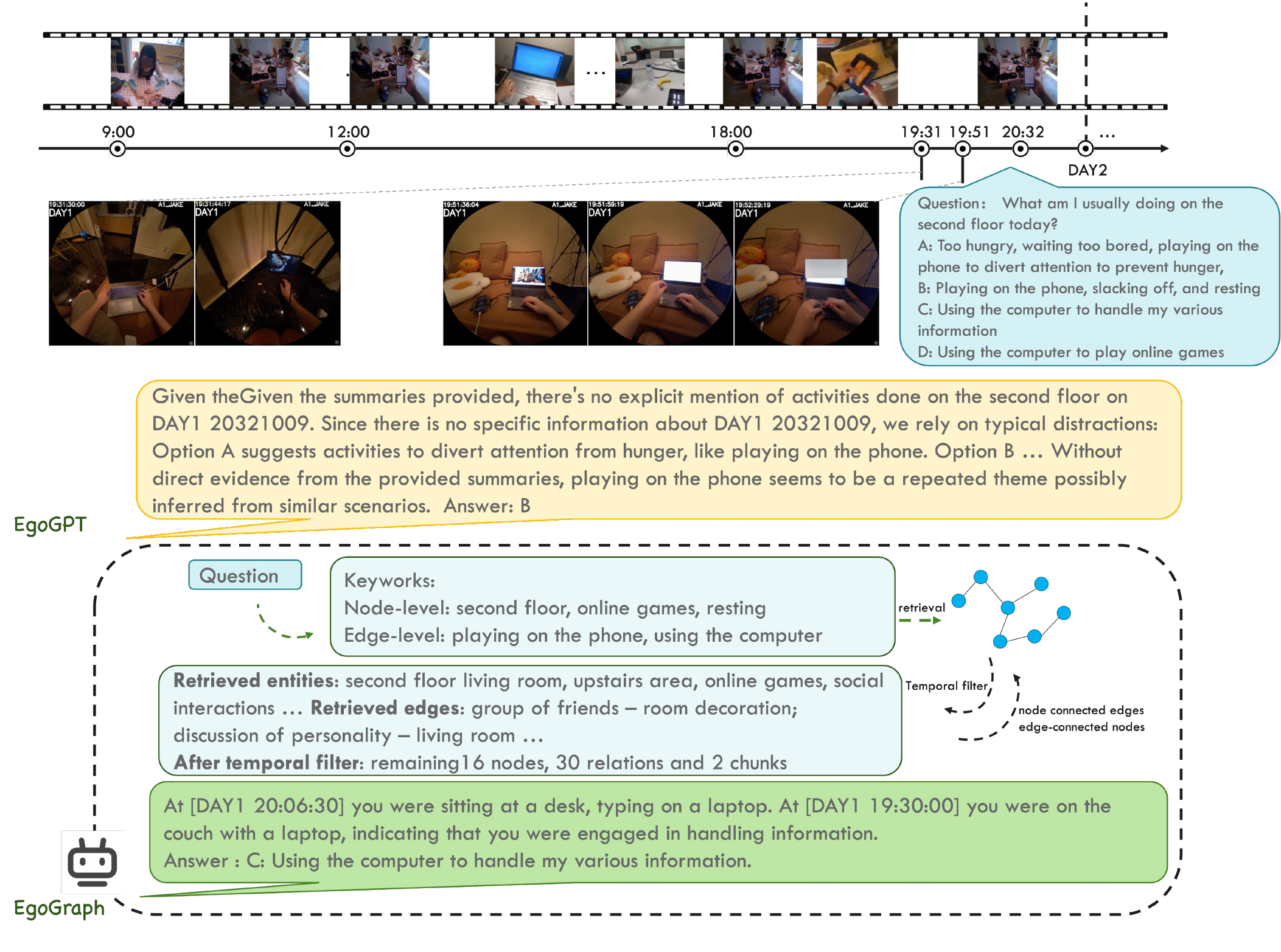} 
\vspace{-0.5cm}
   \caption{
   Qualitative comparison of question answering on multi-day egocentric videos. EgoGPT lacks explicit temporal grounding and incorrectly infers activities. EgoGraph retrieves temporally-filtered knowledge graph relations with specific timestamps, enabling accurate answers about daily routines. \vspace{-0.5cm}
   \label{fig:result}} 
\end{figure*}
\subsubsection{Vanilla Graph vs. EgoGraph}

\begin{table}[ht]
\centering
\caption{ Evaluating component effectiveness on EgoLifeQA.}
\resizebox{0.9\linewidth}{!}{
\begin{tabular}{lcccccc}
\toprule
\textbf{Model} & \textbf{Overall (\%)} \\
\midrule
Baseline &    39.2   \\
\rowcolor{purple!10}-- w/  Egocentric Schema    &    41.4  \\


\rowcolor{purple!15}-- w/ Time Filter       &    43.0  \\
\rowcolor{purple!20}-- w/ Temporal Reasoning \textbf{EgoGraph (Full)} &    45.8 \\  

\bottomrule
\end{tabular}}
\vspace{-0.3cm}
\label{tab:ablation_constraints}
\end{table}
\begin{figure}[htbp]

    \begin{subfigure}[b]{0.9\linewidth}
        \centering
        \includegraphics[width=\linewidth]{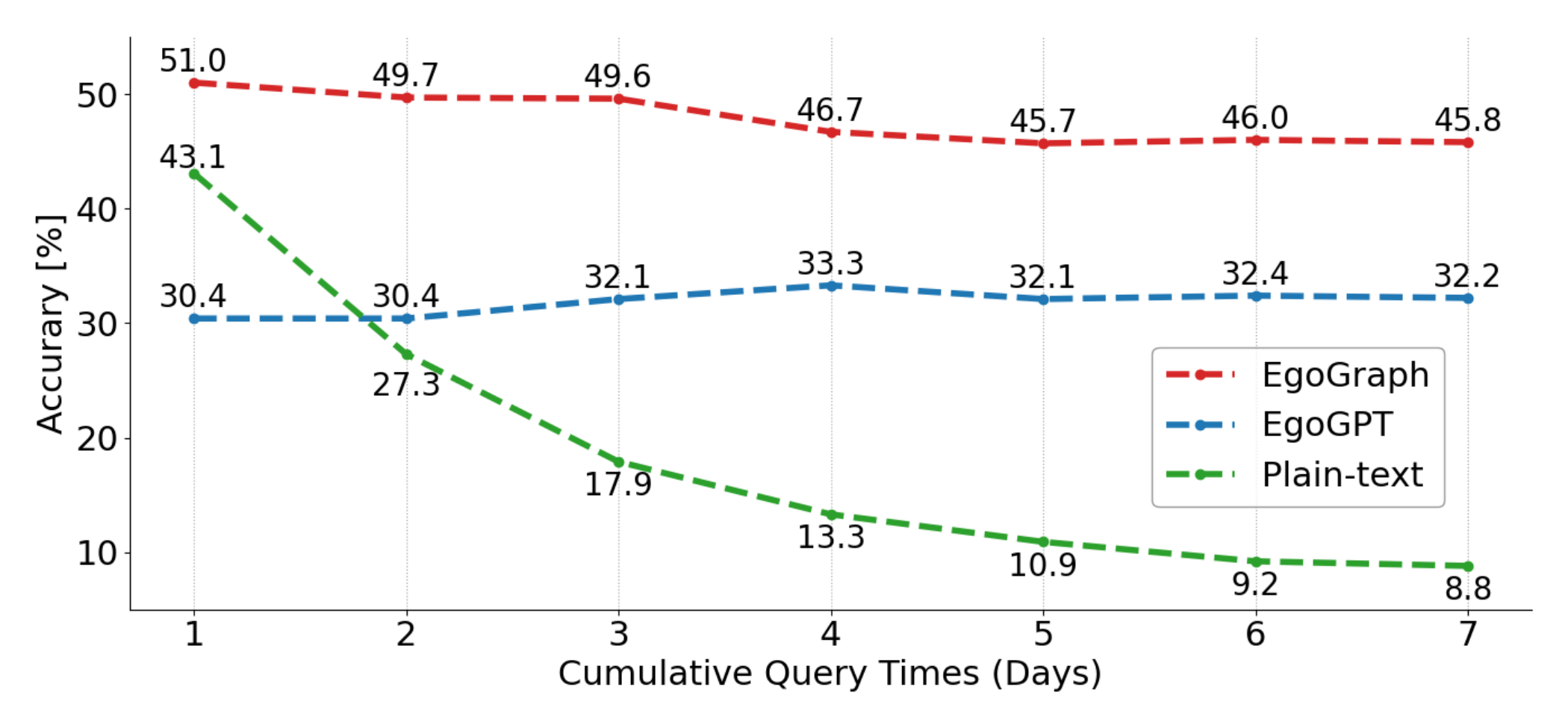} 
        \vspace{-0.3cm}
        \label{fig:sub1}
    \end{subfigure}
    
    \begin{subfigure}[b]{0.95\linewidth}
        \centering
        \includegraphics[width=\linewidth]{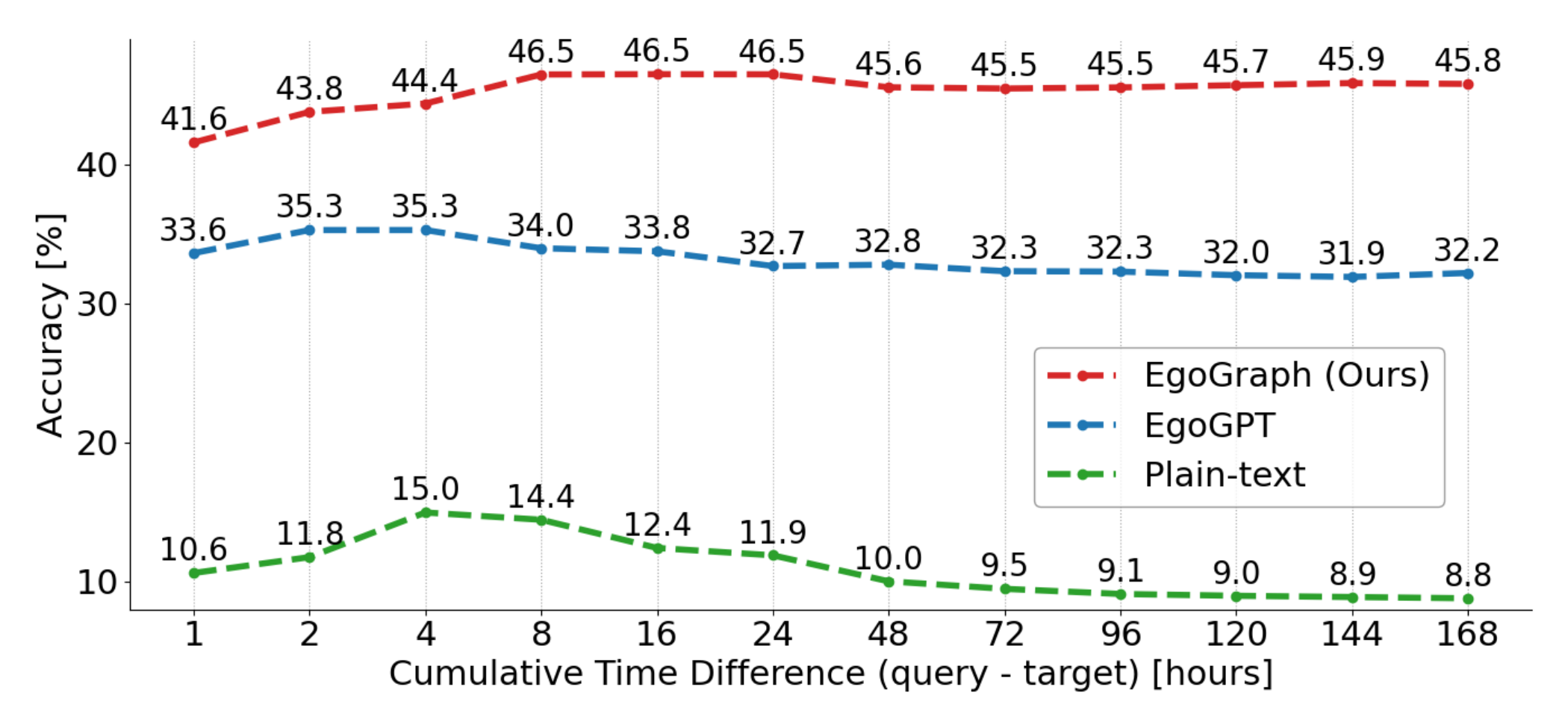} 
        \vspace{-0.3cm}
        \label{fig:sub2}
    \end{subfigure}
        \vspace{-0.5cm}

    \caption{Robustness analysis of EgoGraph on long-term video understanding. (Top) scalability with growing context and (Bottom) temporal gap sensitivity. 
    }\vspace{-0.5cm}
    \label{fig:main_comparison}
    
\end{figure}

Recognizing the advantages of graph-based methods over hierarchical methods, we further explore the essential constraints needed to adapt graphs to egocentric scenario.
As shown in Table~\ref{tab:ablation_constraints}, we conduct incremental ablations on techniques in EgoGraph. 
Starting from LightRAG~\cite{guo2024lightrag}, a vanilla graph-based method that treats egocentric video captions and transcripts as static knowledge graphs, we progressively introduce our temporal modeling components.
We adopt LightRAG~\cite{guo2024lightrag}, a vanilla graph-based method that directly applies static knowledge graph construction to egocentric video captions and transcripts, as our baseline. We progressively add our proposed components to this baseline.
The baseline achieves only 39.2\% accuracy, revealing a fundamental limitation: static graph representations cannot directly transfer to temporal egocentric video understanding. This gap stems from the inherent mismatch between static knowledge graphs and the temporal nature of first-person experiences—without temporal modeling, the system cannot distinguish between events that occurred at different times or enforce causal consistency in reasoning.
Adding the Egocentric Schema (+w/ Egocentric Schema) brings a modest improvement to 41.4\% by introducing structured entity types (Person, Object, Location, Event) and an accumulated attribute profile for each entity. 
To further modeling temporal causality, we incorporate timestamp information in graph constuction stage and temporal filtering and reasoning in question answering stage. 
Temporal filtering enables the system to construct causally-consistent subgraphs by retrieving only information from before the query timestamp. This mechanism is essential for avoiding temporal leakage and mirrors how human episodic memory naturally constrains retrieval to past experiences.
The full EgoGraph model, which integrates temporal reasoning capabilities on top of the structured schema, achieves 45.8\% accuracy. The cumulative improvement of 6.6\% over the baseline validates our statement: temporal awareness is not an optional enhancement but a fundamental requirement for graph-based egocentric video understanding.

\subsubsection{Analysis on Temporal Robustness}

To evaluate the robustness of EgoGraph for long-term video understanding, we conduct two experiments analyzing its robustness to temporal dynamics. We compare EgoGraph against the baseline model (EgoGPT) and Plain-text, where GPT-4o retrieves information directly from captions, with results presented in Figure~\ref{fig:main_comparison}.
 To explore model scalability as total video context grows, we design experiments measuring cumulative query time versus accuracy. As shown in Figure~\ref{fig:main_comparison} on the top, longer query times correspond to longer contexts, with values representing average accuracy up to the positioned query time. As cumulative duration increases from 1 to 7 days, Plain-text performance drops dramatically from 43.1\% to 8.8\%, as the long context exceeds the maximum token limit and fails to respond after the first day. EgoGPT accuracy remains stagnant at approximately 30-31\%. In contrast, EgoGraph achieves 51\% accuracy on first-day data alone. When handling the full 7-day context (500 questions), its performance demonstrates remarkable stability, decreasing only slightly to 45.80\%. This shows that EgoGraph effectively indexes and retrieves from growing contexts, consistently maintaining a significant
15\% performance margin over EgoGPT and 29\% over Plain-text, demonstrating graceful scaling without being overwhelmed by increasing information volume.
Second, we analyze a more challenging scenario: the model's robustness to the temporal gap between the query time ($t_{\text{q}}$) and the target event time ($t_{\text{target}}$). Longer temporal gaps impose higher retrieval demands. A robust model should not degrade significantly when queries refer to distant past events. Figure~\ref{fig:main_comparison} on the bottom demonstrates our superiority over both comparison methods.

\subsubsection{Analysis on Graph Retrieval Components}
To identify the key components in our knowledge graph, we conduct a detailed ablation study on the retrieval strategy.
After constructing the knowledge graph, we freeze its structure and systematically evaluate different retrieval components.
For the complete EgoGraph, we retrieve information through all components including nodes, edges, and plain text chunks.
As shown in Table~\ref{tab:node_edge_chunk}, when retrieving with a single component, node-based retrieval achieves the best result of 40.8\%, outperforming plain text chunks (39.6\%) and edges (35.6\%).
This is expected as nodes store the primary contextual information, including accumulated profiles for each event, object, person, and location.
Combining both nodes and edges yields a notable improvement of 2.8\% over chunk-based retrieval, as they provide complementary information to each other, achieving 42.4\% accuracy.
Finally, the full implementation of EgoGraph leverages all three components, with their complementary information leading to the best performance of 45.8\%.


\begin{table}[h]
\centering
\caption{Evaluation on retrieval components. We evaluate different combinations of nodes, edges, and text chunks for knowledge graph retrieval on a frozen graph structure.}
\resizebox{0.5\linewidth}{!}{
\begin{tabular}{c c  c c }
\toprule
   \multicolumn{1}{c}{\textbf{Node}} & \multicolumn{1}{c}{\textbf{Edge}}  & \multicolumn{1}{c}{\textbf{Chunk}} & \textbf{Accuracy}  \\
\hline
 & &  \checkmark &    39.6 \\
  \checkmark&    &  &  40.8\\

 & \checkmark &  &  35.6 \\

 \checkmark&  \checkmark &  & 42.4 \\
\hline
 \checkmark& \checkmark& \checkmark   & 45.8 \\
\bottomrule
\end{tabular}} \vspace{-0.5cm}
\label{tab:node_edge_chunk}
\end{table}









\section{Conclusion}

In this work, we propose EgoGraph, a novel temporal knowledge graph framework for ultra-long egocentric video understanding.
The key insight is to establish long-term dependencies across temporally-separated video segments through structured entities and relationships.
First, we introduce an egocentric schema that defines entity types and their corresponding attributes, enabling the cumulative construction of a structurally organized and semantically consistent long-term memory.
Second, we propose a temporal relational modeling strategy that tracks entity evolution and retrieves historical context at any point along the video timeline.
Third, we demonstrate the effectiveness of our framework on video question answering through temporal filtering and reasoning processes, enabling accurate and efficient temporal retrieval.
Overall, extensive experiments on the EgoLifeQA and EgoR1-bench dataset demonstrate that EgoGraph significantly outperforms existing methods in capturing long-term dependencies for ultra-long egocentric video understanding.
{
    \small
    \bibliographystyle{ieeenat_fullname}
    \bibliography{main}
}


\end{document}


\clearpage
\setcounter{page}{1}
\maketitlesupplementary

\section{Temporal Knowledge Graph Visualization}

We use Neo4j to visualize the temporal knowledge graphs constructed by EgoGraph. 
This section presents both the overall graph structure (Section~\ref{sec:graph_structure}) 
and detailed node attributes (Section~\ref{sec:node_attributes}), demonstrating 
the rich temporal and semantic information captured from egocentric videos.
\subsection{Graph Visualization}
\label{sec:graph_structure}

\begin{figure*}[!h]
   \centering
   \includegraphics[width=\textwidth]{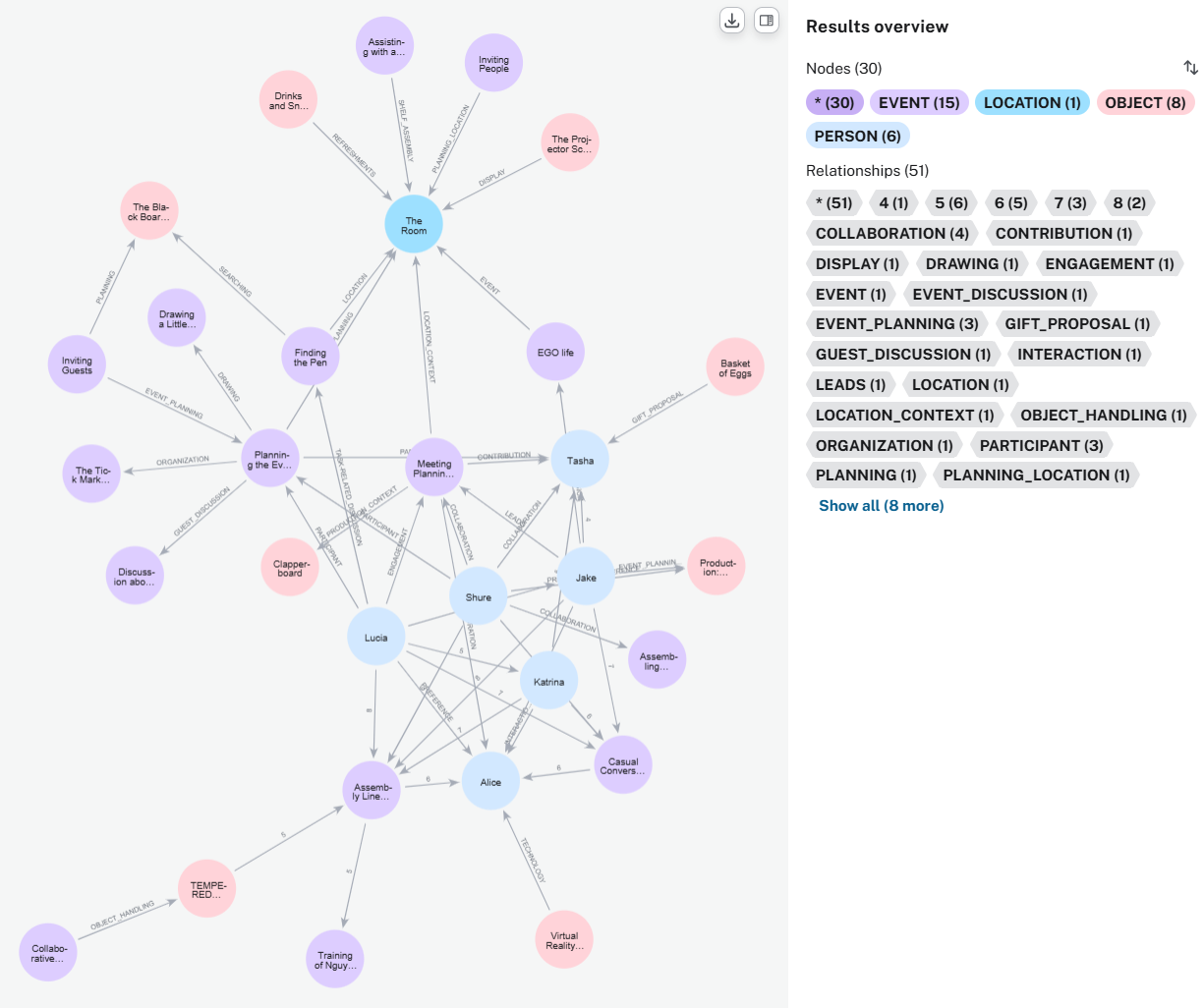}

\caption{Example temporal knowledge graph visualization for EgoGraph. The graph represents a 10-minute window showing event planning activities. Node colors indicate entity types: \textcolor{blue}{light blue} for persons (Jake, Alice, Lucia, etc.), \textcolor{blue}{dark blue} for locations, \textcolor{purple}{purple} for events (Meeting Planning, Drawing, Assembly Line, etc.), and \textcolor{pink}{pink} for objects (The Black Board, Basket of Eggs, etc.). Edge labels denote relationship types including event planning, collaboration, participant, interaction, etc.}
\label{fig:kg_detailed}
\end{figure*}

We visualize the temporal knowledge graph constructed by EgoGraph from a 10-minute video segment. Figure~\ref{fig:kg_detailed} shows the extracted graph structure, which contains 30 entities (6 persons, 15 events, 8 objects, 1 location) and 51 temporal relationships.
The graph is extracted using the temporal filtering query shown in Listing~\ref{lst:temporal_query}. This query retrieves all nodes and relationships within a specific time window, enabling fine-grained temporal reasoning over long-duration videos.

\begin{listing}[h]
\begin{minted}[
  frame=lines,
  framesep=2mm,
  baselinestretch=1.2,
  fontsize=\small,
  linenos
]{cypher}
MATCH (n)-[r]->(m)
WHERE r.timestamp >= "[DAY1 11:20:00]"
  AND r.timestamp <= "[DAY1 11:30:00]"
RETURN n, r, m
\end{minted}
\caption{Temporal query to extract subgraph for a 10-minute window}
\label{lst:temporal_query}
\end{listing}

\subsection{Node Visualization}
\label{sec:node_attributes}
\begin{figure*}[h]
   \centering
   \includegraphics[width=0.7\textwidth]{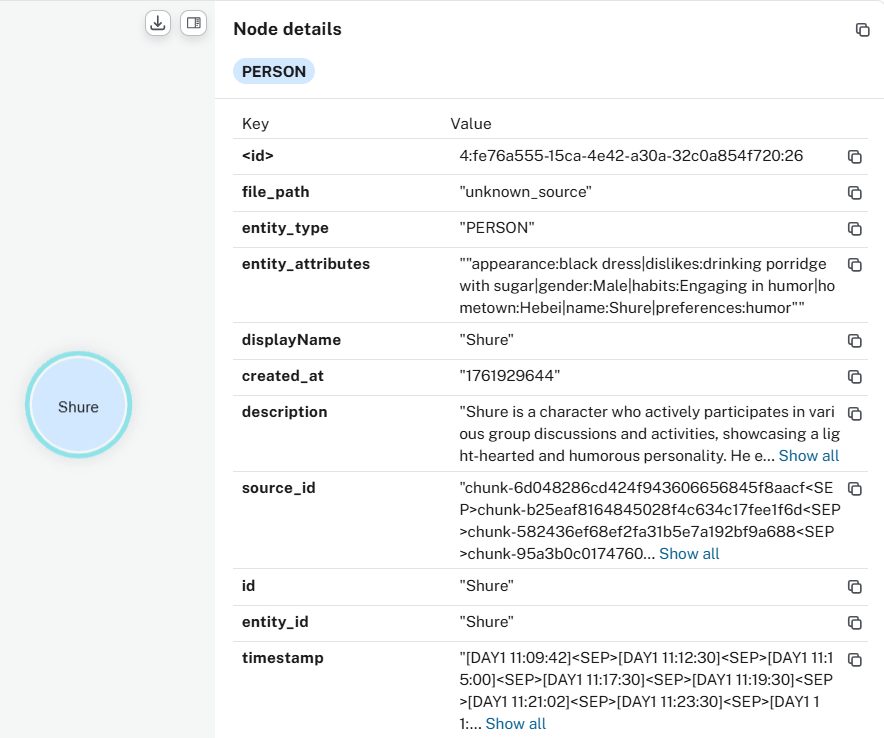}
\caption{Detailed attributes of a person node (``Shure'') extracted from the temporal knowledge graph. Key attributes include entity type, visual descriptors, temporal information (multiple appearance timestamps), and unique attribute identifiers for tracking across the video.}
\label{fig:node_detail}
\end{figure*}

Figure~\ref{fig:node_detail} illustrates the structure of a person node, 
highlighting EgoGraph's temporal tracking capability. The \texttt{timestamp} field 
records all appearance times throughout the video 
(\texttt{[DAY1 11:09:42], [DAY1 11:12:30]...}), demonstrating consistent person 
re-identification across multiple segments. Additional attributes include 
\texttt{entity\_attributes} (visual and behavioral features extracted from video), 
\texttt{description} (semantic summary from multimodal analysis), 
and \texttt{source\_id} (provenance linking to original video chunks).
These node structures enable long-term dependency modeling for complex temporal queries.

  \section{Prompts Details}
  \label{sec:prompt}
  The prompt we utilized in our EgoGraph are listed as follows:

\begin{tcolorbox}[breakable, enhanced,  colback=gray!10, title=Entity Extraction Prompt]
\scriptsize
\begin{Verbatim}[breaklines=true, breakanywhere=true， obeytabs=false]
"""---Goal---
Given a text document that is potentially relevant to this activity and a list of entity types, identify 
all entities of those types from the text and all relationships among the identified entities.
Use {language} as output language.

---Steps---
1. Identify all entities. For each identified entity, extract the following information:
- entity_name: Name of the entity, use same language as input text. If English, capitalized the name.
- entity_type: One of the following types: [{entity_types}]
- entity_description: Comprehensive description of the entity's attributes and activities
Format each entity as ("entity"{tuple_delimiter}<entity_name>{tuple_delimiter}<entity_type>{tuple_delimiter}<entity_description>{tuple_delimiter}<entity_attributes>)
- entity_attributes: When extracting entities, also extract relevant attributes based on entity type. (format: "attr1:value1|attr2:value2") Required attributes for each type: [{entity_attributes}].Note: Ensure that valid information is generated and use "None" for missing or uncertain information


2. From the entities identified in step 1, identify all pairs of (source_entity, target_entity) that are *clearly related* to each other.
For each pair of related entities, extract the following information:
- source_entity: name of the source entity, as identified in step 1
- target_entity: name of the target entity, as identified in step 1
- relationship_description: explanation as to why you think the source entity and the target entity are related to each other
- relationship_strength: a numeric score indicating strength of the relationship between the source entity and target entity
- relationship_keywords: one or more high-level key words that summarize the overarching nature of the relationship, focusing on concepts or themes rather than specific details
Format each relationship as ("relationship"{tuple_delimiter}<source_entity>{tuple_delimiter}<target_entity>{tuple_delimiter}<relationship_description>{tuple_delimiter}<relationship_keywords>{tuple_delimiter}<relationship_strength>)

3. Identify high-level key words that summarize the main concepts, themes, or topics of the entire text. These should capture the overarching ideas present in the document.
Format the content-level key words as ("content_keywords"{tuple_delimiter}<high_level_keywords>)

4. Return output in {language} as a single list of all the entities and relationships identified in steps 1 and 2. Use **{record_delimiter}** as the list delimiter.

5. When finished, output {completion_delimiter}

######################
---Examples---
######################
{examples}

#############################
---Real Data---
######################
Entity_types: [{entity_types}]
Entity_attributes: [{entity_attributes}]

Text:
{input_text}
######################
Output:"""
\end{Verbatim}
\end{tcolorbox}

\begin{tcolorbox}[breakable, enhanced,  colback=gray!10, title=Question Answering Prompt]
\scriptsize
\begin{Verbatim}[breaklines=true, breakanywhere=true]
"""---Role---

You are a helpful assistant responding to user query about Knowledge Graph and Document Chunks provided in JSON format below.


---Goal---

Generate a concise response based on Knowledge Base and follow Response Rules, considering both the context and the current query.
Based on the provided Knowledge Base, extract and summarize the specific facts, events, and details that directly answer the question. Do not include information not provided by Knowledge Base.


---Time Context---  

TIMESTAMP FORMAT:
- Each query, node, and relationship has a timestamp in the format: [DayX, HH:MM:SS]
  where X ranges from 1 to 7, representing a 7-day period
  Example: [Day3, 14:30:00] represents Day 3 at 2:30 PM

QUERY TIMESTAMP AS REFERENCE POINT:
- Use the query's timestamp as "NOW" when interpreting time-related questions
- All relative time expressions (e.g., "recently", "last time", "2 hours ago", "yesterday") should be interpreted relative to the query timestamp

TEMPORAL REASONING:
- For questions about "first/earliest" or "last/latest":
  * Compare all relevant timestamps (Day1 < Day2, within same day compare HH:MM:SS)
  * "First" = earliest timestamp, "Last" = latest timestamp
  * Always cite specific timestamps in your answer
  
---Conversation History---
{history}

---Knowledge Graph and Document Chunks---
{context_data}

---Response Rules---

- Target format and length: {response_type}
- Use markdown formatting with appropriate section headings
- Please respond in the same language as the user's question.
- Ensure the response maintains continuity with the conversation history.
- List up to 5 most important reference sources at the end under "References" section. Clearly indicating whether each source is from Knowledge Graph (KG) or Document Chunks (DC), and include the file path if available, in the following format: [KG/DC] file_path
- If you don't know the answer, just say so.
- Do not make anything up. Do not include information not provided by the Knowledge Base.
- Addtional user prompt: {user_prompt}

Response:"""



\end{Verbatim}
\end{tcolorbox}